\documentclass{article}

\usepackage{iclr2027_conference,times}

\usepackage{amsmath,amssymb,mathtools,array}
\usepackage{graphicx}
\usepackage{booktabs}
\usepackage{multirow}
\usepackage{microtype}
\usepackage{xspace}
\usepackage{hyperref}
\usepackage{url}

\newcommand{\Full}{\textsc{Full}\xspace}
\newcommand{\ExactTopK}{\textsc{Exact Top-K}\xspace}
\newcommand{\PISAZero}{\textsc{PISA-0th}\xspace}
\newcommand{\estci}[3]{\begin{tabular}[c]{@{}c@{}}$#1$\\$[#2,\,#3]$\end{tabular}}
\newcommand{\cellhead}[1]{\shortstack{#1}}
\newsavebox{\papertablebox}
\newcommand{\printpapertable}[1]{\typeout{PAPER-TABLE-WIDTH #1: \the\wd\papertablebox; LIMIT: \the\linewidth}\ifdim\wd\papertablebox>\linewidth\PackageWarning{paper-layout}{Table #1 exceeds linewidth}\fi\usebox{\papertablebox}}

\title{Counterexamples to Local Reconstruction Gain as a Proxy for Final Fidelity in Residual Completion}

\author{Yasuto Hoshi, Daisuke Miyashita \& Jun Deguchi \\
Kioxia Corporation
}

\iclrfinalcopy 

\begin{document}
\maketitle
\lhead{}

\begin{abstract}
	Residual completion augments query-aware sparse attention by estimating the contribution of tokens omitted from the exact sparse computation. We ask whether improving a layer's attention-output reconstruction on the same incoming Q/K/V and selected support necessarily improves the fidelity of the final model output. We study training-free RESA and learned Top-K+$\phi$ with frozen backbone language models. A prespecified single-layer screen yields two Qwen3-0.6B/Multi-LexSum interventions for which direct-runtime measurements show positive prespecified request-aggregate local reconstruction gain but worse final KL fidelity than the corresponding all-abstain \ExactTopK baseline on both discovery and prompt-token-disjoint holdout requests. Exact restoration at the same layer instead improves final fidelity, showing that the reversal is specific to approximate completion in these cases. In complementary multi-layer experiments, a task-independent local diagnostic often repairs the tested completion estimators, although the repaired models do not consistently outperform \ExactTopK. Together, these results show that better local reconstruction need not translate into better final-model fidelity.
\end{abstract}

\section{Introduction}
\label{sec:introduction}

Query-aware sparse attention reduces attention cost by evaluating only a small, query-dependent subset of stored keys and values. Methods such as QUEST and SparQ retain the KV cache while selecting different tokens for each query~\citep{tang2024quest,ribar2024sparq}. Residual completion estimates the aggregate attention contribution of the tokens outside this exact sparse computation and adds that estimate to the sparse attention output~\citep{yang2026resa,hoshi2026residualmass,zhan2026reskv}. Selection therefore determines which tokens are evaluated exactly, while completion approximates the contribution of the remaining tokens.

At a given layer and decode step, completion can be compared with abstention on the same incoming Q/K/V and the same selected support. Our baseline is the corresponding all-abstain \ExactTopK model, which uses the same sparse-selection mechanism but disables the completion branch at every layer. We ask first whether completion makes the attention output closer to dense attention at the selected layer on the measured decode steps; we call this \emph{local reconstruction}. We separately ask whether the resulting model's next-token distribution is closer to that of the dense \Full model; we call this \emph{final fidelity} and measure it using KL divergence to \Full. Later hidden states, queries, and selected supports are allowed to change after the intervention.

Local and final fidelity measure different quantities. After the selected layer, residual blocks and attention layers transform the intervention and may also change subsequent query-dependent token selections. A correction can therefore reduce same-input attention-output error while increasing divergence at the final model output. Prior work also motivates examining downstream consequences of sparse-attention perturbations. Delta Attention links sparse attention-output shift to later query--key misalignment, while RippleKV measures how layerwise KV perturbations affect the final predictive distribution~\citep{willette2025delta,xu2026ripplekv}. Complementary diagnostic and counterfactual studies examine how cache compression and sparse selection alter model behavior~\citep{qiu2026kvdiagnosis,ren2026counterfactual}. We test this local-to-final relationship directly for residual completion.

We study learned Top-K+$\phi$~\citep{hoshi2026residualmass} and training-free RESA~\citep{yang2026resa} on long-context tasks, with additional experiments spanning Llama and Qwen backbones. Both operate on an unchanged, frozen backbone language model: Top-K+$\phi$ trains only the auxiliary completion estimator, while RESA requires no additional training. In a frozen screen of single-layer interventions, five actions show nominal local-positive/final-negative intervals and two survive the prespecified screening multiplicity correction. We fix those two actions and measure them directly at runtime. Completion and abstention runs have identical incoming hidden states, Q/K/V, and selected token IDs at the measured steps, and the attention outputs are those actually returned by the model. On both the discovery requests and a prompt-token-disjoint holdout, the selected interventions improve the prespecified local reconstruction metric while worsening final KL fidelity. These experiments provide concrete counterexamples to the implication that better local attention reconstruction necessarily yields better final-model fidelity.

A separate exact-restoration control replaces sparse attention with dense attention only at the same selected layer, while leaving the other layers sparse. Exact restoration improves final fidelity where approximate completion reduces it, so the observed reversal is not an inevitable consequence of moving the selected layer toward dense attention. The control does not localize the later network component that produces the divergence. A focused post-hoc Qwen3-8B/RESA study provides broader scale context: among 18 layer--task--context actions, one has CI-negative final fidelity despite positive local reconstruction.

Local reconstruction remains useful as a diagnostic in a separate multi-layer study. Using generic, task-independent calibration sequences, we disable completion at layers with negative mean local reconstruction gain. This improves the tested $\phi$/RESA completion models in 11 of 12 fidelity comparisons relative to their unmodified versions, but the repaired models do not consistently outperform \ExactTopK. The same masking rule gives mixed results for \PISAZero~\citep{li2026pisa}, indicating that the diagnostic is estimator-dependent.

Taken together, the experiments show that local reconstruction quality and final-model fidelity are related but not interchangeable endpoints. A completion method can improve one without improving the other, and a layerwise diagnostic can repair an estimator without making the repaired model better than abstention.

\section{Completion and Its Evaluation}
\label{sec:formulation}
\paragraph{Retained cache and selected support.}
We retain the full prefix KV cache. For each query, the exact branch contains four sink tokens, the most recent 64 prompt tokens, exact score-based Top-$K$ tokens from the middle prompt region, and all generated tokens. \ExactTopK normalizes attention over this selected support with completion disabled. We use \emph{abstention} for disabling the completion branch at a layer; the all-abstain \ExactTopK path disables it at every layer. Completion augments the selected-support computation with an estimate of the excluded contribution. In a single-layer intervention, every other layer uses \ExactTopK. The selection rule and budget are unchanged, although later queries and selected token identities may change after the intervention.

\paragraph{Local reconstruction gain.}
At a measured prediction step, let $O_{\mathrm{dense}}$, $O_{\mathrm{TK}}$, and $O_{\mathrm{C}}$ denote the post-$W_O$ dense attention output, Exact Top-K attention output, and attention output with residual completion enabled, respectively, computed from the same incoming Q/K/V. Define
\begin{equation}
	g=1-\frac{\|O_{\mathrm{dense}}-O_{\mathrm{C}}\|^2}
	{\|O_{\mathrm{dense}}-O_{\mathrm{TK}}\|^2}.
	\label{eq:gain}
\end{equation}
Thus $g>0$ exactly when completion is closer to the dense attention output than \ExactTopK is under squared reconstruction error. When $|O_{\mathrm{dense}}-O_{\mathrm{TK}}|$ is above the validity threshold, abstention has $g=0$ by construction; when this residual is too small, normalized gain is left undefined rather than set to zero. We aggregate in two stages. For request $i$, we first compute $G_i=\operatorname{median}_{t\in S_i}g_{it}$ over predetermined measured steps $S_i$. The reported local reconstruction gain is then the request-level mean, $G=\frac{1}{N}\sum_i G_i$. Absolute squared-error reduction and alternative local aggregates are reported separately. Appendix~\ref{app:runtime-details} gives the validity rule and aggregation details.

\paragraph{Final-model fidelity.}
All compared models receive the same prompt and dense-model-generated continuation tokens. If $T_i$ is the set of evaluated prediction steps, the request-level final-fidelity gain is
\begin{equation}
	F_i=\frac{1}{|T_i|}\sum_{t\in T_i}\left[
	D_{\mathrm{KL}}(p^{\mathrm{Full}}_{it}\Vert p^{\mathrm{TK}}_{it})
	-D_{\mathrm{KL}}(p^{\mathrm{Full}}_{it}\Vert p^{\mathrm{C}}_{it})\right].
	\label{eq:finalgain}
\end{equation}
Here $p^{\mathrm{TK}}$ denotes the corresponding all-abstain \ExactTopK path. Positive $F=\frac{1}{N}\sum_i F_i$ favors completion over abstention; negative $F$ favors abstention. The primary $F$ uses all evaluated decode predictions. A separate $F_S$ restricts Eq.~\ref{eq:finalgain} to $S_i$, aligning the final evaluation with the steps sampled for local measurement. Free-running task utility, $U$, is measured independently.

\section{Experimental Design}
\label{sec:design}
\paragraph{Models, estimators, and tasks.}
The main study includes Llama-3.2-1B/3B-Instruct and Qwen3-0.6B/1.7B~\citep{grattafiori2024llama3,yang2025qwen3} with a learned positive-feature estimator, $\phi$, and the training-free RESA prior~\citep{hoshi2026residualmass,yang2026resa}. All backbone language-model parameters are held fixed across \Full, \ExactTopK, and completion variants; Top-K+$\phi$ differs only by its auxiliary completion estimator, whereas RESA adds no training. We use frozen estimator configurations throughout. The stress tasks are RULER NIAH Multikey-2 and FWE, and HELMET Multi-LexSum~\citep{hsieh2024ruler,yen2025helmet,shen2022multilexsum}. They were selected using only the \Full--\ExactTopK performance gap, without reference to completion outcomes. The main setting is a 65,536-token prompt and $K=16$, giving 84 $(=4+64+16)$ prompt tokens in the exact branch before generated tokens are added. The $K\in\{8,16,32\}$ range is an aggressive fixed-read stress regime with a large omitted set. It still includes high-accuracy sparse cases: the no-headroom control matches \Full across all tested models and budgets, and other conditions retain substantial task accuracy (Appendix~\ref{app:taskfreeze}). Appendix~\ref{app:config} gives the attention, estimator, and positional settings; Appendix~\ref{app:phi-provenance} documents the frozen $\phi$ checkpoint recipe. A focused post-hoc Qwen3-8B/RESA experiment is reported separately in Appendix~\ref{app:scale8b}.

\paragraph{Screening and fixed interventions.}
For each Qwen model--estimator pair, we select three early/middle/late layers from layers with positive pre- and post-$W_O$ local reconstruction gain on the task-independent generic calibration data described below. Crossing them with three tasks gives 36 single-layer actions, each evaluated on 50 requests. Screening local reconstruction gain is obtained from dense-model replay, while screening final-fidelity gain is obtained from the single-layer sparse intervention. The screening analysis uses a screening-specific \ExactTopK implementation as its baseline and yields two actions after the prespecified screening multiplicity correction: $\phi$ L15 and RESA L23, both Qwen3-0.6B/Multi-LexSum. These action identities are fixed before the direct-runtime follow-up. The direct-runtime stage uses the estimator-specific \ExactTopK implementation for both completion and abstention. For RESA, the screening-specific and estimator-specific baselines give slightly different baseline KL values while the completion KL is unchanged; consequently, the reported effect size changes with the operational baseline. We therefore treat screening and direct-runtime as distinct analysis stages rather than repeated estimates of one numerically identical effect, and we do not use the screening estimate as runtime evidence (Appendix~\ref{app:baseline}). The holdout construction removes candidates with the same tokenized prompt as a discovery request, as well as duplicate candidates, and retains the first 50 eligible prompts in source order; holdout outcomes are not used to choose the fixed actions. The discovery and holdout request sets are then fixed for direct-runtime evaluation. Appendix~\ref{app:onelayer} reports the complete screen and the fixed candidate layers, and Appendix~\ref{app:holdout} details the holdout construction.

\paragraph{Direct-runtime measurement.}
For each fixed action and request set, we execute \Full, the single-layer completion intervention, and the all-abstain \ExactTopK baseline using the same estimator-specific sparse implementation. Prefill uses dense attention; the first, prefill-produced prediction is excluded from the decode comparison. The local measurement samples up to 32 prediction steps at predetermined, evenly spaced positions. At those steps, the \ExactTopK baseline and completion intervention have bitwise-equal incoming hidden states, post-rotary queries, full cached keys/values, and selected token IDs. We observe the outputs actually returned by the attention modules, rather than recomputing completion offline. The local dense reference uses FP32 softmax on the observed Q/K/V and the output projection in the model's native dtype. For the first two requests of each direct-runtime evaluation, all evaluated full-vocabulary logits are also compared with measurement disabled to check non-interference. Appendix~\ref{app:runtime-details} specifies the measurement and aggregation protocol; Appendix~\ref{app:baseline} explains the baseline difference between the screening and estimator-specific direct-runtime comparisons.

\paragraph{Exact-restoration control.}
A separate control replaces the selected layer's sparse attention by dense attention, leaving the other layers sparse. We compare approximate completion and exact restoration with the same estimator-specific all-abstain \ExactTopK baseline. The block-state metric $G_{\mathrm{block}}$ measures relative squared error to the global \Full hidden state, while $F$ remains the final-fidelity gain relative to \ExactTopK. Both differ from the same-input local reference in Eq.~\ref{eq:gain}; in particular, $G_{\mathrm{block}}$ is not the local reconstruction gain $G$ defined above. This control tests the downstream effect of exact restoration at the selected layer. Appendix~\ref{app:absolute-kl} gives the corresponding absolute and relative KL effect sizes.

\paragraph{Task-independent masking.}
For each model, 30 unlabeled 64K calibration sequences are drawn equally from FineWeb, arXiv Summarization, and BIGPATENT~\citep{penedo2024fineweb,cohan2018arxivsum,sharma2019bigpatent}. We freeze the set of layers satisfying
\begin{equation}
	\overline G_{\mathrm{generic},\ell}<0
	\label{eq:selector}
\end{equation}
where $\overline G_{\mathrm{generic},\ell}$ is the mean post-$W_O$ local reconstruction gain for layer $\ell$ over these calibration sequences. This defines the task-independent negative-$G$ masking rule: completion is disabled at the selected layers. The repaired hybrid is compared with both unmodified completion and all-abstain \ExactTopK. Same-count anti-ranked masks and ten depth-matched random masks provide alternative-mask controls. We also apply the rule to an adapted zeroth-order PISA estimator~\citep{li2026pisa} to test its dependence on the residual representation. Appendix~\ref{app:repair-controls} reports the condition-level masking, alternative-mask, and calibration-stability controls; Appendix~\ref{app:pisa} gives the \PISAZero transfer results.

\paragraph{Statistical units and multiplicity.}
Uncertainty is computed by paired resampling of requests after within-request aggregation. The 36-action screen applies one-sided percentile bounds to 72 directional statements at $\alpha/72$, with $\alpha=0.05$. The direct-runtime follow-up is a separate four-direction family per split, using 100,000 bootstrap draws and cutoff $\alpha/4$. Tables otherwise report paired 95\% intervals; sensitivity endpoints are analyzed separately.

\section{Results}
\label{sec:results}
\subsection{Direct-runtime local improvement and final degradation}
\begin{figure}[t]
	\centering
	\includegraphics[width=\linewidth]{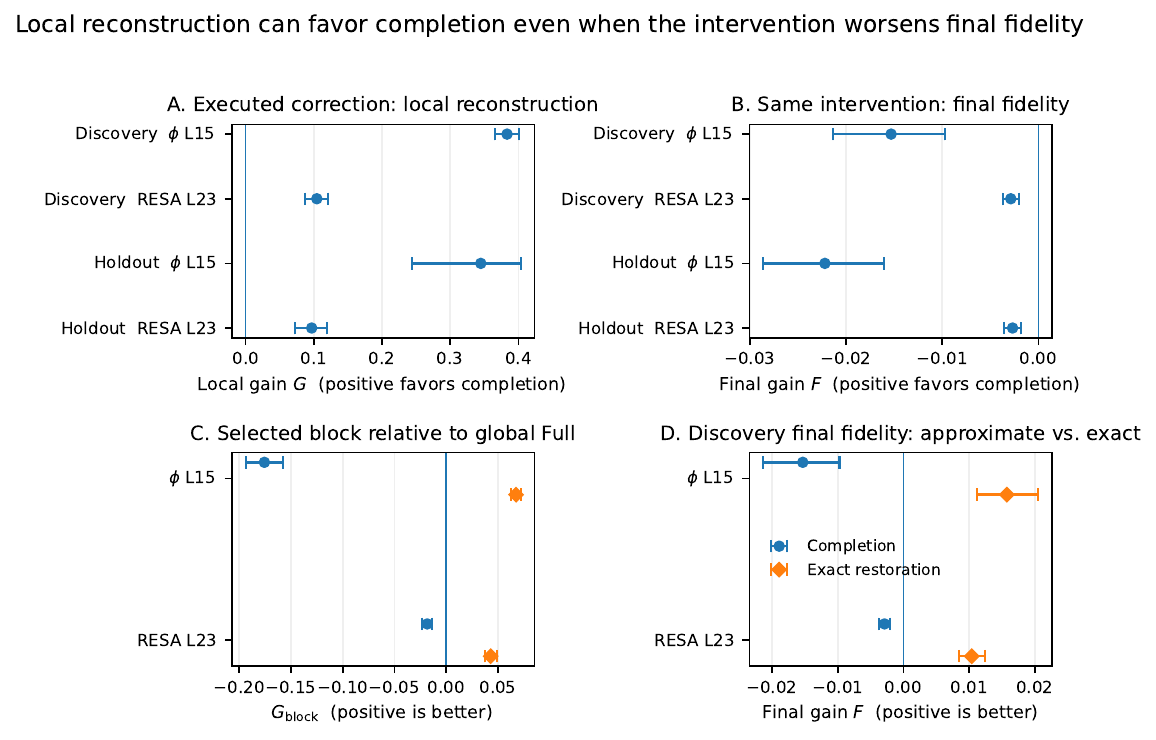}
	\caption{\textbf{Direct-runtime mismatch and exact-restoration control.} Panels A--B show the same-input local reconstruction gain $G$ and final-fidelity gain $F$ for the two fixed actions on discovery and holdout requests. Positive $G$ but negative $F$ is the primary mismatch. Panels C--D show the separate discovery-set exact-restoration control at the \Full-relative block and final endpoints. The block metric $G_{\mathrm{block}}$ in this control compares hidden-state error to the global \Full trajectory and is not the same-input local $G$ in Panel A. Approximate completion is worse than \ExactTopK at the block and final endpoints, whereas exact restoration at the same selected layer is better. Error bars are paired 95\% intervals. Because the local and block metrics use different reference states, the figure does not localize where the mismatch arises.}
	\label{fig:runtime-story}
\end{figure}

The 36-action screen contains 26 actions with CI-positive local reconstruction gain. Of these, 5 also have CI-negative final-fidelity gain, and 2 survive the 72-direction screening multiplicity correction. Appendix Figure~\ref{fig:discovery} reports the complete screen and documents selection rather than the direct-runtime premise. The two selected actions are then measured directly in Table~\ref{tab:runtime-core} and summarized visually in Figure~\ref{fig:runtime-story}.

\begin{table}[t]
	\centering
	\small
	\setlength{\tabcolsep}{4.0pt}
	\renewcommand{\arraystretch}{1.12}
	\caption{\textbf{Local reconstruction gain and final-fidelity gain measured in the same execution.} Both actions are Qwen3-0.6B/Multi-LexSum at 64K/$K=16$, with 50 requests per split. \emph{Discovery} denotes the 50-request screening pool used to identify the fixed interventions. \emph{Holdout} denotes a 50-request pool constructed after removing candidates with the same tokenized prompt as a discovery request and duplicate candidates; its outcomes were not used for action selection (Appendix~\ref{app:holdout}). $G$ averages within-request medians over sampled steps; $F$ averages all evaluated prediction steps. Entries give request-level means and paired 95\% intervals; for $G$, each request-level value is the median over its sampled steps. The last column counts observed request summaries with $G_i>0,F_i<0$, not individually significant effects.}
	\label{tab:runtime-core}
	\sbox{\papertablebox}{%
		\begin{tabular}{@{}lcccc@{}}
			\toprule
			Split     & Action     & Local reconstruction $G\uparrow$ & Final-fidelity $F\uparrow$        & \cellhead{Requests \\$G_i>0,F_i<0$} \\
			\midrule
			Discovery & $\phi$ L15 & \estci{0.383}{0.365}{0.401}      & \estci{-0.0153}{-0.0213}{-0.0098} & $41/50$            \\
			Discovery & RESA L23   & \estci{0.104}{0.087}{0.121}      & \estci{-0.0029}{-0.0037}{-0.0020} & $43/50$            \\
			Holdout   & $\phi$ L15 & \estci{0.345}{0.244}{0.404}      & \estci{-0.0222}{-0.0286}{-0.0160} & $43/50$            \\
			Holdout   & RESA L23   & \estci{0.097}{0.072}{0.119}      & \estci{-0.0027}{-0.0036}{-0.0018} & $38/50$            \\
			\bottomrule
		\end{tabular}}
	\printpapertable{tab:runtime-core}
\end{table}

Under the prespecified median-based $G$, both interventions have positive local reconstruction gain while all-step final-fidelity gain is negative relative to the all-abstain \ExactTopK baseline on both request sets. All four action--split comparisons retain these directions under the four-direction sensitivity applied separately to each split. The joint counts show that the same ordering also occurs within individual request summaries, rather than only in the marginal means.

Alternative local-reconstruction summaries are less uniform. In particular, on the $\phi$ holdout, row-mean $G$, energy-pooled $G$, and absolute local error reduction all have intervals crossing zero. The holdout counterexample therefore applies to the prespecified within-request-median criterion rather than to every plausible local-error aggregation; Appendix~\ref{app:runtime-sensitivity} reports the full sensitivity results.

The three nominal-only FWE cases do not show CI-negative final-fidelity gain on their prompt-token-disjoint holdouts (Appendix~\ref{app:holdout}). Captured-state replay at $K=8,16,32$ preserves the local-positive/final-negative ordering for the two core actions within the same stress regime (Appendix~\ref{app:k-sensitivity}).

\subsection{Exact restoration and approximate completion have opposite effects}
Figure~\ref{fig:runtime-story} and Table~\ref{tab:core-mechanism} compare approximate completion with exact restoration on the separate discovery control. This comparison asks a different question from the same-input local reconstruction in Table~\ref{tab:runtime-core}: $G_{\mathrm{block}}$ follows the intervention trajectory and compares its selected-block hidden state with the global \Full trajectory, so its sign need not match the local $G$. At both the selected-block and final endpoints, approximate completion is worse than \ExactTopK and exact restoration is better. In this comparison, exact restoration reduces final KL by about 4.7\% for $\phi$ L15 and 3.1\% for RESA L23, whereas approximate completion increases it by about 4.6\% and 0.9\%, respectively. Absolute KL and paired relative changes are reported in Appendix~\ref{app:absolute-kl}.

\begin{table}[t]
	\centering
	\small
	\setlength{\tabcolsep}{4.0pt}
	\renewcommand{\arraystretch}{1.12}
	\caption{\textbf{Approximate completion versus exact restoration.} Separate discovery-set controls use the same estimator-specific all-abstain \ExactTopK baseline. $G_{\mathrm{block}}$ compares hidden-state error to the global \Full trajectory at the selected block and is not the same-input local $G$ reported in Table~\ref{tab:runtime-core}; $F$ is final-fidelity gain relative to the same \ExactTopK baseline. Values are means and paired 95\% intervals.}
	\label{tab:core-mechanism}
	\sbox{\papertablebox}{%
		\begin{tabular}{@{}lccc@{}}
			\toprule
			Action     & Intervention      & $G_{\mathrm{block}}\uparrow$   & Final-fidelity $F\uparrow$        \\
			\midrule
			$\phi$ L15 & Completion        & \estci{-0.176}{-0.194}{-0.158} & \estci{-0.0153}{-0.0213}{-0.0097} \\
			$\phi$ L15 & Exact restoration & \estci{0.067}{0.063}{0.072}    & \estci{0.0157}{0.0111}{0.0206}    \\
			RESA L23   & Completion        & \estci{-0.018}{-0.023}{-0.014} & \estci{-0.0029}{-0.0037}{-0.0020} \\
			RESA L23   & Exact restoration & \estci{0.043}{0.037}{0.049}    & \estci{0.0104}{0.0085}{0.0124}    \\
			\bottomrule
		\end{tabular}}
	\printpapertable{tab:core-mechanism}
\end{table}

Exact restoration therefore has the opposite block- and final-level effect from approximate completion in these cases. Because the local and block metrics use different reference states, this control distinguishes the interventions but does not localize a unique downstream location where the mismatch arises.

\subsection{Repairing completion does not establish an advantage over \ExactTopK}
The negative-$G$ masking rule selects no layers in the four Llama model--estimator pairs (Appendix Table~\ref{tab:generic-calibration}). It selects 9 and 12 layers for Qwen3-0.6B $\phi$ and RESA, and 5 and 12 for Qwen3-1.7B. Disabling these sets improves final fidelity relative to the unmodified completion model in 11/12 treated conditions and utility in 8/12. Against all-abstain \ExactTopK, however, the repaired hybrids give 3 positive, 4 unresolved, and 5 negative fidelity comparisons (Table~\ref{tab:repair}). The corresponding utility counts are 1 positive, 9 unresolved, and 2 negative.

\begin{table}[t]
	\centering
	\small
	\setlength{\tabcolsep}{4.0pt}
	\renewcommand{\arraystretch}{1.12}
	\caption{\textbf{Repair and comparison with \ExactTopK.} \emph{Off} is the number of layers with completion disabled. Counts are positive / unresolved / negative paired 95\% intervals over three tasks. Llama rows are absent because the selector disables no layers in the four Llama model--estimator pairs. Repair compares the masked hybrid with unmodified completion; vs. TK compares it with the all-abstain \ExactTopK baseline. $F$: final-fidelity gain; $U$: task-utility gain relative to the stated reference.}
	\label{tab:repair}
	\sbox{\papertablebox}{%
		\begin{tabular}{@{}lccccc@{}}
			\toprule
			Model / est.        & Off  & \cellhead{Repair                               \\$F$} & \cellhead{Repair\\$U$} & \cellhead{vs. TK\\$F$} & \cellhead{vs. TK\\$U$} \\
			\midrule
			Qwen3-0.6B / $\phi$ & $9$  & $2/1/0$          & $1/2/0$ & $2/1/0$ & $1/2/0$ \\
			Qwen3-0.6B / RESA   & $12$ & $3/0/0$          & $2/1/0$ & $0/1/2$ & $0/2/1$ \\
			Qwen3-1.7B / $\phi$ & $5$  & $3/0/0$          & $2/1/0$ & $1/1/1$ & $0/2/1$ \\
			Qwen3-1.7B / RESA   & $12$ & $3/0/0$          & $3/0/0$ & $0/1/2$ & $0/3/0$ \\
			\bottomrule
		\end{tabular}}
	\printpapertable{tab:repair}
\end{table}

The selected sets outperform the same-count anti-ranked masks in all 12 fidelity comparisons, and their point-estimate repairs exceed every one of the ten fixed depth-matched masks in each condition (Appendix~\ref{app:repair-controls}). Yet repair and advantage over the all-abstain \ExactTopK baseline remain different questions. For example, Qwen3-0.6B/RESA on Multi-LexSum gains $0.733$ in final fidelity relative to unmodified completion, but the repaired hybrid is still $0.076$ worse than \ExactTopK. A large repair can therefore leave the model worse than the all-abstain \ExactTopK baseline.

The diagnostic is also estimator-dependent. Applying the same negative-$G$ masking rule to \PISAZero yields 2 positive, 5 unresolved, and 5 negative fidelity repairs (Appendix~\ref{app:pisa}). Local reconstruction gain can therefore be useful for locating estimator-specific failure, but its sign alone does not determine the final-model ordering.

For broader context, Appendix~\ref{app:stage-endpoints} reports observational all-layer completion endpoints across Llama and Qwen. All 12 Llama conditions have CI-positive local and final-fidelity gains, whereas eight Qwen conditions combine CI-positive post-$W_O$ local gain with CI-negative final fidelity. Appendix~\ref{app:scale8b} further reports a focused Qwen3-8B/RESA study: 7 of 18 actions are CI-positive in final fidelity, 10 are unresolved, and 1 is CI-negative. These broader results show that the mismatch is condition-dependent and are not additional replications of the prespecified direct-runtime core.

\section{Related Work}
\label{sec:related}
\paragraph{Query-aware selection and residual estimation.}
QUEST and SparQ reduce attention work by selecting query-dependent subsets of the retained KV cache, while MagicPIG estimates attention through sampling rather than deterministic Top-$K$~\citep{tang2024quest,ribar2024sparq,chen2025magicpig}. Residual-estimation methods such as RESA, Top-K+$\phi$, and ResKV instead approximate the contribution omitted by a sparse branch~\citep{yang2026resa,hoshi2026residualmass,zhan2026reskv}. These works primarily evaluate the resulting sparse-attention method; our experiment isolates a single completion intervention and asks how its realized local reconstruction error relates to the final model output.

\paragraph{Cache compression, eviction, and reconstruction.}
Cache-compression methods alter a different part of the inference state: H$_2$O and InfiniGen change which KV content is retained or fetched, while LESS adds an auxiliary recurrent representation~\citep{zhang2023h2o,lee2024infinigen,dong2024less}. ReST-KV is particularly close to our question because it uses layer-wise output-reconstruction discrepancy to guide KV eviction~\citep{an2026restkv}. Our setting keeps the full prefix cache and the exact selected support available, so the intervention is confined to the completion output rather than cache retention. This lets us examine whether improved local reconstruction preserves the ordering at the final model output.

\paragraph{Propagation and final-output sensitivity.}
Delta Attention studies downstream effects of sparse attention-output shift, while RippleKV measures the response of the final predictive distribution to layerwise KV perturbations~\citep{willette2025delta,xu2026ripplekv}. KVDiagnosis analyzes compression failures using cache-, attention-, likelihood-, and decoding-level diagnostics, while counterfactual sparse-attention evaluation measures how sparsification changes the influence of selected content on model outputs~\citep{qiu2026kvdiagnosis,ren2026counterfactual}. Our experiments connect these downstream and diagnostic perspectives to residual completion while holding the retained cache and exact-support rule fixed.

\section{Discussion}
The direct-runtime experiments show a sign reversal between same-input local reconstruction and final KL fidelity for two fixed interventions, and the reversal is reproduced on prompt-token-disjoint holdout requests. The result is strongest for the prespecified within-request-median $G$; alternative local aggregations are less stable, especially for the $\phi$ holdout. The evidence therefore establishes concrete local-to-final counterexamples without estimating how often they occur.

The exact-restoration control distinguishes approximate completion from exact correction. At the same selected layer, dense restoration improves final fidelity while approximate completion reduces it. The divergence therefore cannot be explained simply by the model being harmed whenever that layer is moved toward its dense-attention output; the responsible downstream transformation remains unresolved.

The masking study gives a complementary view of the same issue. Negative-$G$ masks improve final fidelity in 11 of 12 tested $\phi$/RESA conditions relative to their unmodified versions, but some repaired models remain worse than the all-abstain \ExactTopK baseline, and the same rule does not transfer uniformly to \PISAZero. Local reconstruction is therefore informative about estimator behavior, but it is not a substitute for measuring the resulting model-level fidelity and utility.

\section{Limitations}
\label{sec:limitations}
\paragraph{Experimental coverage.}
The direct-runtime core contains two selected Qwen3-0.6B/Multi-LexSum interventions at 64K/$K=16$. Its two 50-request pools are prompt-token-disjoint but not established to be source-document- or legal-case-disjoint. The $K=8,16,32$ sensitivity covers nearby aggressive budgets, not substantially denser support. The 8B focused experiment is post-hoc, and Llama/Qwen comparisons also differ in long-context positional settings.

\paragraph{Metrics and interpretation.}
The primary local-reconstruction endpoint averages within-request medians over sampled decode steps, whereas final-fidelity gain $F$ averages all evaluated steps. Alternative local weightings are less uniform: row-mean $G$ is CI-positive in two of four direct comparisons, energy-pooled $G$ in three of four, and absolute local error reduction in three of four; all three are unresolved in the $\phi$ holdout. The central result is therefore specific to the prespecified primary criterion. Final-fidelity gain $F$ measures change in KL fidelity to \Full relative to \ExactTopK rather than task utility, and no targeted one-layer utility loss is statistically resolved. Later support can change after the intervention, and the experiments do not isolate a unique downstream mechanism.

\paragraph{Estimators and systems.}
The study evaluates the frozen estimator configurations documented in Appendix~\ref{app:config}. \PISAZero is an adapted estimator-family control with a different auxiliary representation. Exact score-based Top-$K$ isolates selection/completion behavior; end-to-end retrieval accuracy, latency, bandwidth, and serving-optimal budgets are outside the measured endpoints.

\section{Conclusion}
For two fixed residual-completion interventions, positive prespecified local reconstruction gain coexists with worse final dense-model fidelity on both discovery and holdout requests. Exact restoration at the same layer produces the opposite final effect, and negative-$G$ masking often repairs broader completion models without guaranteeing an advantage over \ExactTopK. Local attention reconstruction is therefore useful evidence about an estimator, but it does not by itself predict the fidelity of the final model output.

\subsection*{Reproducibility statement}
The supplement reports the full screening panel, estimator settings, task-selection criteria, fixed layer masks, paired request identities, and uncertainty calculations. Direct-runtime observations and budget-sensitivity replay measurements use their respective protocols and baseline implementations. Deterministic bootstrap streams and request-level outputs support reconstruction of the reported tables without additional model inference.

\subsection*{AI use statement}
Generative AI tools assisted with translation, literature search and summarization, manuscript editing, and analysis-code development. The authors are responsible for the final content and validation of these materials.

\bibliography{iclr2027_kvcache}
\bibliographystyle{iclr2027_conference}
\clearpage
\appendix
\setcounter{table}{0}
\renewcommand{\thetable}{A\arabic{table}}
\renewcommand{\theHtable}{A\arabic{table}}
\setcounter{figure}{0}
\renewcommand{\thefigure}{A\arabic{figure}}
\section{Configuration and Measurement Protocol}
\label{app:config}
\subsection{Attention, positional settings, and estimators}
The full prompt KV cache is retained in all sparse variants. The exact support consists of four initial tokens, 64 recent prompt tokens, the Top-$K$ middle prompt tokens by exact attention score, and all generated tokens. At $K=8,16,32$, this gives 76, 84, and 100 exact prompt positions, respectively. These values intentionally vary the middle selection within an aggressive fixed-read stress regime, leaving a large omitted set rather than spanning deployment-optimal budgets. The regime does not imply task failure: NIAH Single-1 matches \Full for all four backbones at all three budgets, and several other conditions retain substantial task accuracy (Table~\ref{tab:taskfreeze}). Llama uses its native positional configuration; Qwen3 uses YaRN factor 2 to extend the 32,768-token base context to 65,536 tokens~\citep{peng2024yarn}. Within-model comparisons use the fixed evaluation configuration. Numerical differences between the screening-specific \ExactTopK baseline and the RESA-specific \ExactTopK baseline are described in Appendix~\ref{app:baseline}.

The two main residual estimators are Top-K+$\phi$ and RESA. Top-K+$\phi$ uses a positive-feature representation of omitted attention numerator and normalizer terms. RESA obtains a typical-query prior at the end of prefill and combines that prior with the selected sparse branch during decode~\citep{hoshi2026residualmass,yang2026resa}. The experiments use these frozen configurations rather than training or tuning an estimator on downstream outcomes. \PISAZero is an adapted zeroth-order residual estimator: the omitted middle region is partitioned into fixed blocks, each block is represented by one zeroth-order summary, and those summaries provide the completion branch alongside the exact selected-token branch. The \PISAZero experiments use 64-token blocks.

\subsection{Frozen $\phi$ checkpoint recipe}
\label{app:phi-provenance}
Each backbone uses a frozen 200,000-step checkpoint trained with 65,536-token prefill packs and bfloat16 teachers. FineWeb, arXiv Summarization, and BIGPATENT are scheduled equally in round-robin order~\citep{penedo2024fineweb,cohan2018arxivsum,sharma2019bigpatent}. The ReZero-$\phi$ map has 512-dimensional embeddings, 64-dimensional features, one MLP layer, and a positive exponential activation. There are 64 stratified query rows per pack, with weights 0.15, 0.25, and 0.60 assigned to positions 0--4095, 4096--16383, and the final 33K tokens. Loss weights are $\lambda_{\mathrm{KL}}=0.99$ and $\lambda_{\mathrm{top}}=0.1$; false-positive, log-normalizer, and attention-output loss weights are zero. This is a deliberate simplification of the training objective in~\citet{hoshi2026residualmass}. Their residual-denominator diagnostics report underestimation as the dominant pattern in three of four corpus/length settings, while also showing that the sign is not universal; their formulation notes that underestimation is already constrained indirectly by the KL and top-band terms. Motivated by that observation, these checkpoints retain temperature-scaled KL and top-band Huber shaping while omitting the false-positive and one-sided log-normalizer penalties; the separate attention-output auxiliary loss in our training implementation is also disabled. Residual-aware training is disabled and $K_{\mathrm{train}}=0$.

Teacher-only calibration sets a model-global temperature $\tau$ by inverting the corpus-balanced normalized-attention-entropy curve at $H^*=0.9995$. For logit depth $d_{qi}=\max_j s_{qj}-s_{qi}$, a head-specific band $\Delta_{\ell h}$ targets corpus-balanced coverage $c^*=0.95$. These H9995g/c95h settings are included in the checkpoint metadata. Qwen uses YaRN factor 2 during this training recipe; Llama retains its native positional configuration.

\subsection{Direct observations and uncertainty}
\label{app:runtime-details}
For each of the two fixed actions and each fixed request set, three models process the same prompt and teacher tokens: dense \Full, the estimator-specific all-abstain \ExactTopK path, and its single-layer completion path. We record the actual input to, and output from, the selected attention module. At up to 32 evenly spaced decode positions per request, incoming hidden states, post-RoPE Q, the full K/V tensors returned by cache update, and actual selected token IDs must match exactly between the two sparse paths. Dense reference attention is evaluated on those tensors with FP32 softmax, followed by the output projection in the model's native dtype. The observed completion and abstention outputs are used directly rather than replaced by recomputed approximations.

For local residual energy below the validity floor,
\[
	\|O_{\mathrm{dense}}-O_{\mathrm{TK}}\|
	\leq \max\!\left(\epsilon,10^{-6}\|O_{\mathrm{dense}}\|\right),
\]
normalized gain is undefined rather than set to zero. Absolute squared-error reduction,
$\|O_{\mathrm{dense}}-O_{\mathrm{TK}}\|^2-\|O_{\mathrm{dense}}-O_{\mathrm{C}}\|^2$,
does not divide by residual energy. The primary local statistic takes a median within request; supplementary row-mean and energy-pooled statistics use different weightings. All uncertainty estimates resample requests, not heads or decode rows as independent samples.

For the first two requests of each direct-runtime evaluation, baseline and action are also executed with measurement disabled. Equality of all evaluated full-vocabulary logits checks that instrumentation does not change the forward output on these requests. This check is distinct from the input/support comparisons made at sampled local positions throughout the pool. The direct-runtime family has two actions and two directions per existing split. Ordinary intervals use the 2.5th and 97.5th percentiles; directional bounds use $\alpha/4$ and $1-\alpha/4$ for $\alpha=0.05$. The screening family contains 72 directional statements. Bootstrap streams are deterministic and keyed by the statistic identity.

\clearpage
\section{Direct-runtime Sensitivities}
\label{app:runtime-sensitivity}
The primary local-reconstruction endpoint is $G$, a mean across requests of within-request median normalized gains. Table~\ref{tab:runtime-sensitivity} additionally reports row-mean and energy-pooled normalized gains, absolute local squared-error reduction, and $F_S$, which evaluates final-fidelity gain only at the locally measured prediction steps. These estimates complement the principal all-step $F$ comparison; their ordinary intervals are not additional multiplicity-corrected discoveries.
\begin{table}[t]
	\centering
	\small
	\setlength{\tabcolsep}{2.6pt}
	\renewcommand{\arraystretch}{1.12}
	\caption{Direct-runtime sensitivity endpoints. Row-mean and energy-pooled $G$ reweight the same local measurements relative to the primary within-request median. Absolute error reduction is
	$\|O_{\mathrm{dense}}-O_{\mathrm{TK}}\|^2-\|O_{\mathrm{dense}}-O_{\mathrm{C}}\|^2$,
	averaged within and across requests. $F_S$ uses exactly the prediction steps measured for local reconstruction gain. These are ordinary paired 95\% intervals; the primary four-direction analysis uses median $G$ and all-step $F$.}
	\label{tab:runtime-sensitivity}
	\sbox{\papertablebox}{%
		\begin{tabular}{@{}lccccc@{}}
			\toprule
			Split     & Action     & Row-mean $G\uparrow$         & Energy-pooled $G\uparrow$     & \cellhead{Absolute local                                                      \\error reduction $\uparrow$} & Sampled-step $F_S\uparrow$ \\
			\midrule
			Discovery & $\phi$ L15 & \estci{0.194}{0.156}{0.229}  & \estci{0.082}{0.038}{0.125}   & \estci{1.076}{0.572}{1.572}    & \estci{-0.0121}{-0.0245}{4.264\times10^{-5}} \\
			Discovery & RESA L23   & \estci{0.085}{0.059}{0.109}  & \estci{0.100}{0.077}{0.121}   & \estci{45.703}{34.832}{56.646} & \estci{-0.0035}{-0.0058}{-0.0014}            \\
			Holdout   & $\phi$ L15 & \estci{0.095}{-0.047}{0.183} & \estci{-0.003}{-0.126}{0.079} & \estci{0.237}{-0.952}{1.116}   & \estci{-0.0229}{-0.0342}{-0.0123}            \\
			Holdout   & RESA L23   & \estci{0.026}{-0.049}{0.086} & \estci{0.085}{0.051}{0.114}   & \estci{41.583}{28.560}{53.845} & \estci{-0.0036}{-0.0062}{-0.0010}            \\
			\bottomrule
		\end{tabular}}
	\printpapertable{tab:runtime-sensitivity}
\end{table}

The four-direction bounds for the primary local-reconstruction/final-fidelity family are Discovery $\phi$ L15: $L_G=0.363$, $U_F=-0.0090$; Discovery RESA L23: $L_G=0.085$, $U_F=-0.0019$; Holdout $\phi$ L15: $L_G=0.214$, $U_F=-0.0152$; Holdout RESA L23: $L_G=0.068$, $U_F=-0.0017$.

The same request weighting is used within each endpoint, but the endpoints differ in normalization and step coverage. Uncertainty in a sensitivity endpoint does not inherit the directional conclusion from the primary comparison.

\clearpage
\section{Screening, Task Selection, and Replay Comparisons}
\label{app:onelayer}
\subsection{Task selection}
\label{app:taskfreeze}
Only \Full and \ExactTopK outcomes enter stress-task selection. Eligible tasks retain positive headroom for all four backbones at 64K/$K=16$ and all 12 model--budget points over $K\in\{8,16,32\}$. The three eligible tasks with the largest mean gap are fixed. NIAH Single-1 serves as a no-headroom control. Table~\ref{tab:taskfreeze} includes the full ranking, including tasks that do not meet the criterion.
\begin{table}[t]
	\centering
	\small
	\setlength{\tabcolsep}{4.0pt}
	\renewcommand{\arraystretch}{1.12}
	\caption{Task-selection ranking. Gaps are \Full minus \ExactTopK in percentage points, averaged or minimized across four models. The last columns count positive gaps out of four models and 12 model--budget points. Ranks 1--3 are the selected stress tasks; NIAH Single-1 is the no-headroom control.}
	\label{tab:taskfreeze}
	\sbox{\papertablebox}{%
		\begin{tabular}{@{}lccccc@{}}
			\toprule
			Rank & Task            & \cellhead{Mean                             \\gap} & \cellhead{Min.\\gap} & $K=16$ $+$ & All-$K$ $+$ \\
			\midrule
			$1$  & NIAH Multikey-2 & $30.5$         & $18.0$  & $4/4$ & $12/12$ \\
			$2$  & FWE             & $20.25$        & $7.3$   & $4/4$ & $12/12$ \\
			$3$  & Multi-LexSum    & $6.36$         & $4.92$  & $4/4$ & $12/12$ \\
			$4$  & NIAH Multiquery & $6.33$         & $3.2$   & $4/4$ & $12/12$ \\
			$5$  & HELMET RAG      & $2.58$         & $1.59$  & $4/4$ & $12/12$ \\
			$6$  & NIAH Single-3   & $10.75$        & $0.0$   & $3/4$ & $9/12$  \\
			$7$  & NIAH Multikey-3 & $8.0$          & $0.0$   & $3/4$ & $9/12$  \\
			$8$  & NIAH Multivalue & $5.47$         & $-2.7$  & $3/4$ & $8/12$  \\
			$9$  & NarrativeQA     & $4.83$         & $-0.34$ & $3/4$ & $11/12$ \\
			$10$ & QA SQuAD        & $3.5$          & $-5.2$  & $3/4$ & $9/12$  \\
			$11$ & VT              & $3.25$         & $-9.8$  & $3/4$ & $8/12$  \\
			$12$ & NIAH Multikey-1 & $2.25$         & $0.0$   & $3/4$ & $8/12$  \\
			$13$ & QA Hotpot       & $1.5$          & $-3.0$  & $2/4$ & $7/12$  \\
			$14$ & NIAH Single-2   & $0.25$         & $-1.0$  & $1/4$ & $4/12$  \\
			$15$ & NIAH Single-1   & $0.0$          & $0.0$   & $0/4$ & $0/12$  \\
			$16$ & CWE             & $-0.73$        & $-2.5$  & $2/4$ & $6/12$  \\
			\bottomrule
		\end{tabular}}
	\printpapertable{tab:taskfreeze}
\end{table}

\subsection{All candidate layers and screening endpoints}
The screening panel crosses two Qwen backbones, two estimators, three candidate layers with positive generic-calibration local reconstruction gain, and three tasks. All candidates appear in Table~\ref{tab:onelayer}. Screening local reconstruction gain uses dense-model replay, and screening final-fidelity gain uses the screening baseline. Five actions have nominal local-positive/final-negative intervals, and two survive the 72-direction screening multiplicity correction. Direct-runtime results for those two actions appear in the main text with estimator-specific \ExactTopK baselines.
\begin{table}[t]
	\centering
	\small
	\setlength{\tabcolsep}{4.0pt}
	\renewcommand{\arraystretch}{1.12}
	\caption{All candidate layers in the 36-action screen, each tested on three tasks. A mismatch task has both a nominal CI-positive local reconstruction gain and a nominal CI-negative final-fidelity gain. An asterisk identifies an action that survives the 72-direction screening multiplicity correction. Local reconstruction gain uses dense-model replay; final-fidelity gain uses the screening baseline.}
	\label{tab:onelayer}
	\sbox{\papertablebox}{%
		\begin{tabular}{@{}lccccc@{}}
			\toprule
			Model      & Est.   & Layer & Local $G$ CI$+$ & Final-fidelity $F$ CI$-$ & Mismatch task         \\
			\midrule
			Qwen3-0.6B & $\phi$ & $6$   & $3/3$           & $1/3$                    & FWE                   \\
			Qwen3-0.6B & $\phi$ & $15$  & $1/3$           & $1/3$                    & Multi-LexSum$^{\ast}$ \\
			Qwen3-0.6B & $\phi$ & $23$  & $2/3$           & $1/3$                    & \textemdash           \\
			Qwen3-0.6B & RESA   & $6$   & $3/3$           & $0/3$                    & \textemdash           \\
			Qwen3-0.6B & RESA   & $12$  & $3/3$           & $1/3$                    & FWE                   \\
			Qwen3-0.6B & RESA   & $23$  & $2/3$           & $2/3$                    & Multi-LexSum$^{\ast}$ \\
			Qwen3-1.7B & $\phi$ & $6$   & $2/3$           & $1/3$                    & \textemdash           \\
			Qwen3-1.7B & $\phi$ & $12$  & $1/3$           & $2/3$                    & \textemdash           \\
			Qwen3-1.7B & $\phi$ & $23$  & $1/3$           & $1/3$                    & \textemdash           \\
			Qwen3-1.7B & RESA   & $6$   & $3/3$           & $0/3$                    & \textemdash           \\
			Qwen3-1.7B & RESA   & $12$  & $3/3$           & $1/3$                    & FWE                   \\
			Qwen3-1.7B & RESA   & $22$  & $2/3$           & $1/3$                    & \textemdash           \\
			\bottomrule
		\end{tabular}}
	\printpapertable{tab:onelayer}
\end{table}

\begin{figure}[t]
	\centering
	\includegraphics[width=.88\linewidth]{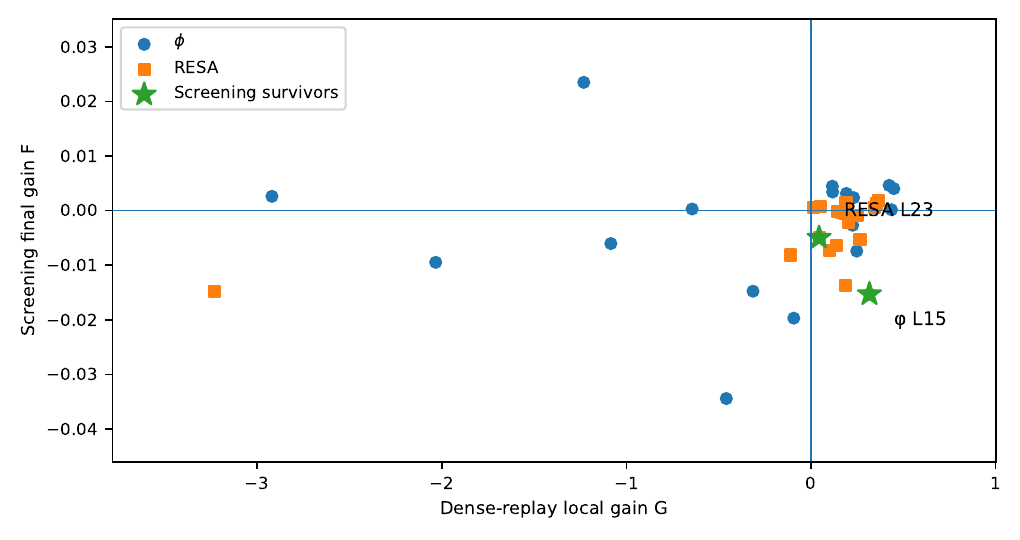}
	\caption{\textbf{Complete 36-action discovery screen.} Each point is a model--estimator--layer--task action. Local reconstruction gain comes from dense-model replay; final-fidelity gain $F$ uses the screening baseline. Stars mark the two actions surviving the 72-direction screening multiplicity correction. This figure documents the selection stage; the main direct-runtime evidence is Figure~\ref{fig:runtime-story} and Table~\ref{tab:runtime-core}.}
	\label{fig:discovery}
\end{figure}

\subsection{Prompt-token-disjoint holdouts under the screening protocol}
\label{app:holdout}
The three remaining nominal cases are fixed before their FWE holdout outcomes are inspected. Discovery prompt-token identities and duplicate candidates are removed, and the first 50 eligible prompts are retained in source order. Table~\ref{tab:holdout} reports all three cases. Captured-state local reconstruction gain remains CI-positive, but none has a CI-negative final-fidelity gain under the screening protocol. This provides a negative holdout result and is not used as evidence for the direct-runtime core.
\begin{table}[t]
	\centering
	\small
	\setlength{\tabcolsep}{4.0pt}
	\renewcommand{\arraystretch}{1.12}
	\caption{Nominal-only FWE hypotheses on 50 prompt-token-disjoint holdout requests. Model sizes denote Qwen3. The six-direction family-wise bounds $L_G$ and $U_F$ are the lower bound for local reconstruction gain and the upper bound for final-fidelity gain. None has a CI-negative final-fidelity gain.}
	\label{tab:holdout}
	\sbox{\papertablebox}{%
		\begin{tabular}{@{}lcccc@{}}
			\toprule
			Action           & Captured-state $G$          & Screening $F$                       & $L_G$   & $U_F$     \\
			\midrule
			0.6B / $\phi$ L6 & \estci{0.222}{0.184}{0.262} & \estci{0.00257}{-0.00154}{0.00684}  & $0.177$ & $0.00774$ \\
			0.6B / RESA L12  & \estci{0.255}{0.226}{0.286} & \estci{-0.00127}{-0.00849}{0.00582} & $0.219$ & $0.00726$ \\
			1.7B / RESA L12  & \estci{0.331}{0.309}{0.351} & \estci{-0.00733}{-0.02123}{0.00612} & $0.304$ & $0.00902$ \\
			\bottomrule
		\end{tabular}}
	\printpapertable{tab:holdout}
\end{table}

\subsection{Nearby $K$ values}
\label{app:k-sensitivity}
Table~\ref{tab:k-sensitivity} keeps the two core action identities fixed and varies $K\in\{8,16,32\}$. These results combine captured-state local reconstruction gain with screening-protocol final-fidelity gain. The $K=16$ entries correspond to the screening-protocol measurements. All six ordinary local-reconstruction/final-fidelity intervals retain the stated ordering, with smaller RESA final-fidelity effects as $K$ increases. The experiment tests robustness within the deliberately aggressive fixed-read stress regime; it is not a sweep over deployment-optimal or substantially denser supports.
\begin{table}[t]
	\centering
	\small
	\setlength{\tabcolsep}{4.0pt}
	\renewcommand{\arraystretch}{1.12}
	\caption{Nearby-budget sensitivity for the two fixed Qwen3-0.6B/Multi-LexSum actions. Local reconstruction gain uses captured-state replay and final-fidelity gain uses the screening protocol. Each entry gives a mean and ordinary 95\% interval; these six rows are not a new discovery family.}
	\label{tab:k-sensitivity}
	\sbox{\papertablebox}{%
		\begin{tabular}{@{}lccc@{}}
			\toprule
			Action     & $K$  & Captured-state $G$             & Screening $F$                     \\
			\midrule
			RESA L23   & $8$  & \estci{0.1216}{0.1015}{0.1418} & \estci{-0.0103}{-0.0154}{-0.0051} \\
			RESA L23   & $16$ & \estci{0.1079}{0.0884}{0.1269} & \estci{-0.005}{-0.0068}{-0.0032}  \\
			RESA L23   & $32$ & \estci{0.0852}{0.0628}{0.1049} & \estci{-0.0019}{-0.0033}{-0.0005} \\
			$\phi$ L15 & $8$  & \estci{0.4024}{0.3764}{0.4269} & \estci{-0.0127}{-0.0218}{-0.0047} \\
			$\phi$ L15 & $16$ & \estci{0.3852}{0.3556}{0.4118} & \estci{-0.0153}{-0.0212}{-0.0097} \\
			$\phi$ L15 & $32$ & \estci{0.3668}{0.3243}{0.4018} & \estci{-0.011}{-0.015}{-0.0072}   \\
			\bottomrule
		\end{tabular}}
	\printpapertable{tab:k-sensitivity}
\end{table}

\clearpage
\section{Exact Restoration and Baseline-specific Effect Sizes}
\label{app:absolute-kl}
The exact-restoration experiment uses three paths: the all-abstain \ExactTopK baseline, approximate completion at one selected layer, and exact restoration with dense attention at that layer. The other layers remain sparse. The estimator-specific \ExactTopK baseline uses the same sparse backend as each intervention. Block error is measured relative to the global \Full hidden state; it is not the same-input dense-attention reference used for local reconstruction gain. This control uses the same prefix-level consistency check as the corresponding intervention comparison; the main direct-runtime experiment additionally verifies input and support equality between completion and abstention.

The signed final-fidelity gain $F$ measures an absolute KL difference. We also report
\begin{equation}
	R_{\mathrm{KL}}=100\left(\frac{\overline D_{\mathrm{KL}}(p^{\mathrm{Full}}\Vert p^{\mathrm{C}})}
		{\overline D_{\mathrm{KL}}(p^{\mathrm{Full}}\Vert p^{\mathrm{TK}})}-1\right)\%,
	\label{eq:relativekl}
\end{equation}
recomputing both means over the same resampled requests in each bootstrap draw. Positive $R_{\mathrm{KL}}$ means greater divergence from \Full. Table~\ref{tab:absolute-kl-control} gives these effect sizes for the exact-restoration experiment.
\begin{table}[t]
	\centering
	\small
	\setlength{\tabcolsep}{4.0pt}
	\renewcommand{\arraystretch}{1.12}
	\caption{KL effect sizes from the exact-restoration experiment, using the estimator-specific \ExactTopK baseline. Positive relative KL change means greater divergence from \Full; negative means smaller divergence. Means and ordinary paired 95\% intervals are shown.}
	\label{tab:absolute-kl-control}
	\sbox{\papertablebox}{%
		\begin{tabular}{@{}lcccc@{}}
			\toprule
			Action     & Intervention      & Baseline KL                    & Intervention KL                & $R_{\mathrm{KL}}$          \\
			\midrule
			$\phi$ L15 & Completion        & \estci{0.3329}{0.3083}{0.3595} & \estci{0.3482}{0.3236}{0.3751} & \estci{4.6\%}{2.9}{6.5}    \\
			$\phi$ L15 & Exact restoration & \estci{0.3329}{0.3083}{0.3595} & \estci{0.3172}{0.2937}{0.3436} & \estci{-4.7\%}{-6.1}{-3.4} \\
			RESA L23   & Completion        & \estci{0.3350}{0.3104}{0.3621} & \estci{0.3379}{0.3130}{0.3650} & \estci{0.9\%}{0.6}{1.1}    \\
			RESA L23   & Exact restoration & \estci{0.3350}{0.3104}{0.3621} & \estci{0.3247}{0.3001}{0.3515} & \estci{-3.1\%}{-3.7}{-2.5} \\
			\bottomrule
		\end{tabular}}
	\printpapertable{tab:absolute-kl-control}
\end{table}

\subsection{Screening and direct-runtime \ExactTopK baselines}
\label{app:baseline}
For RESA L23, the screening and direct-runtime analyses use slightly different operational implementations of \ExactTopK. In the discovery comparison, the mean baseline KL is $0.332939$ with the screening-specific implementation and $0.335045$ with the estimator-specific implementation, while the completion KL is $0.337919$ in both comparisons. Accordingly, the reported final-fidelity gain changes from approximately $-0.004980$ to $-0.002874$. Because the completion KL is identical in the two analyses, this numerical shift comes from the baseline path. The aggregate outputs alone do not identify a more specific implementation-level cause. We therefore keep screening and direct-runtime effect sizes stage-specific and use the estimator-specific \ExactTopK baseline for all primary direct-runtime comparisons on both request sets.

\clearpage
\section{Calibration and Masking Controls}
\label{app:repair-controls}
\subsection{Condition-level repair and comparison with \ExactTopK}
The task-independent negative-$G$ masking rule disables the complete set of layers with negative generic mean post-$W_O$ local reconstruction gain. It is a set-level intervention, not an assertion that each selected layer is individually harmful. Table~\ref{tab:repair-condition} gives repair relative to unmodified completion and the same-count anti-ranked control. Table~\ref{tab:hybrid-condition} gives the repaired hybrid's direct comparison with \ExactTopK. This separates the reference models behind the two comparisons in the main table.
\begin{table}[t]
	\centering
	\small
	\setlength{\tabcolsep}{4.0pt}
	\renewcommand{\arraystretch}{1.12}
	\caption{Condition-level negative-$G$ masking results. Repair compares with unmodified completion; the last column compares the selected mask with its same-count anti-ranked control. MLX: Multi-LexSum; MK2: NIAH Multikey-2.}
	\label{tab:repair-condition}
	\sbox{\papertablebox}{%
		\begin{tabular}{@{}lccc@{}}
			\toprule
			Task & Fidelity repair              & Utility repair                & \cellhead{Fidelity vs.      \\anti-ranked mask} \\
			\midrule
			\multicolumn{4}{@{}l}{\textit{Qwen3-0.6B / $\phi$; 9 layers disabled}}                            \\
			FWE  & \estci{0.034}{0.025}{0.043}  & \estci{-0.020}{-0.080}{0.040} & \estci{0.033}{0.022}{0.046} \\
			MLX  & \estci{0.005}{-0.009}{0.020} & \estci{1.178}{-2.737}{5.228}  & \estci{0.025}{0.013}{0.037} \\
			MK2  & \estci{0.144}{0.120}{0.170}  & \estci{0.120}{0.020}{0.220}   & \estci{0.163}{0.139}{0.188} \\
			\addlinespace[3pt]
			\multicolumn{4}{@{}l}{\textit{Qwen3-0.6B / RESA; 12 layers disabled}}                             \\
			FWE  & \estci{0.176}{0.140}{0.213}  & \estci{0.047}{-0.013}{0.113}  & \estci{0.133}{0.099}{0.168} \\
			MLX  & \estci{0.733}{0.643}{0.831}  & \estci{13.725}{7.801}{19.942} & \estci{0.590}{0.512}{0.672} \\
			MK2  & \estci{0.257}{0.208}{0.312}  & \estci{0.120}{0.040}{0.220}   & \estci{0.228}{0.178}{0.284} \\
			\addlinespace[3pt]
			\multicolumn{4}{@{}l}{\textit{Qwen3-1.7B / $\phi$; 5 layers disabled}}                            \\
			FWE  & \estci{0.362}{0.286}{0.436}  & \estci{0.153}{0.073}{0.233}   & \estci{0.327}{0.256}{0.398} \\
			MLX  & \estci{0.099}{0.081}{0.119}  & \estci{4.516}{-1.302}{10.208} & \estci{0.102}{0.083}{0.122} \\
			MK2  & \estci{0.083}{0.059}{0.109}  & \estci{0.220}{0.100}{0.340}   & \estci{0.083}{0.060}{0.107} \\
			\addlinespace[3pt]
			\multicolumn{4}{@{}l}{\textit{Qwen3-1.7B / RESA; 12 layers disabled}}                             \\
			FWE  & \estci{0.225}{0.184}{0.269}  & \estci{0.127}{0.047}{0.207}   & \estci{0.188}{0.150}{0.229} \\
			MLX  & \estci{0.449}{0.398}{0.502}  & \estci{9.832}{3.723}{16.027}  & \estci{0.387}{0.341}{0.435} \\
			MK2  & \estci{0.256}{0.224}{0.290}  & \estci{0.380}{0.240}{0.520}   & \estci{0.203}{0.173}{0.233} \\
			\bottomrule
		\end{tabular}}
	\printpapertable{tab:repair-condition}
\end{table}

\label{app:hybrid-vs-topk}
\begin{table}[t]
	\centering
	\small
	\setlength{\tabcolsep}{4.0pt}
	\renewcommand{\arraystretch}{1.12}
	\caption{Repaired hybrid versus the all-abstain \ExactTopK baseline. These paired comparisons have a different reference model from repair versus unmodified completion. MLX: Multi-LexSum; MK2: NIAH Multikey-2.}
	\label{tab:hybrid-condition}
	\sbox{\papertablebox}{%
		\begin{tabular}{@{}lcc@{}}
			\toprule
			Task & Fidelity vs. \ExactTopK        & Utility vs. \ExactTopK         \\
			\midrule
			\multicolumn{3}{@{}l}{\textit{Qwen3-0.6B / $\phi$}}                    \\
			FWE  & \estci{-0.001}{-0.012}{0.010}  & \estci{0.080}{0.007}{0.160}    \\
			MLX  & \estci{0.107}{0.091}{0.125}    & \estci{1.798}{-3.844}{7.530}   \\
			MK2  & \estci{0.039}{0.026}{0.052}    & \estci{-0.060}{-0.180}{0.060}  \\
			\addlinespace[3pt]
			\multicolumn{3}{@{}l}{\textit{Qwen3-0.6B / RESA}}                      \\
			FWE  & \estci{-0.028}{-0.043}{-0.014} & \estci{0.020}{-0.047}{0.087}   \\
			MLX  & \estci{-0.076}{-0.093}{-0.059} & \estci{-4.807}{-9.401}{-0.216} \\
			MK2  & \estci{-0.012}{-0.029}{0.005}  & \estci{-0.080}{-0.180}{0.000}  \\
			\addlinespace[3pt]
			\multicolumn{3}{@{}l}{\textit{Qwen3-1.7B / $\phi$}}                    \\
			FWE  & \estci{0.009}{-0.017}{0.038}   & \estci{0.020}{-0.040}{0.073}   \\
			MLX  & \estci{0.106}{0.081}{0.133}    & \estci{1.966}{-2.471}{6.332}   \\
			MK2  & \estci{-0.039}{-0.054}{-0.025} & \estci{-0.240}{-0.360}{-0.120} \\
			\addlinespace[3pt]
			\multicolumn{3}{@{}l}{\textit{Qwen3-1.7B / RESA}}                      \\
			FWE  & \estci{-0.019}{-0.042}{0.003}  & \estci{0.000}{-0.053}{0.053}   \\
			MLX  & \estci{-0.044}{-0.058}{-0.029} & \estci{-2.191}{-6.611}{1.962}  \\
			MK2  & \estci{-0.028}{-0.036}{-0.021} & \estci{-0.080}{-0.200}{0.040}  \\
			\bottomrule
		\end{tabular}}
	\printpapertable{tab:hybrid-condition}
\end{table}

\subsection{Same-count, depth-matched alternatives}
\label{app:random}
Ten fixed depth-matched masks are evaluated per treated Qwen model--estimator pair. Table~\ref{tab:random} reports the observed range of final-fidelity repairs and the number of masks with repair at least as large as the selected mask. These ranges are not confidence intervals, and the count is not a calibrated $p$-value.
\begin{table}[t]
	\centering
	\small
	\setlength{\tabcolsep}{4.0pt}
	\renewcommand{\arraystretch}{1.12}
	\caption{Same-count, depth-matched mask controls. Ranges summarize ten fixed masks; the last column is an observed count, not a $p$-value. MLX: Multi-LexSum; MK2: NIAH Multikey-2.}
	\label{tab:random}
	\sbox{\papertablebox}{%
		\begin{tabular}{@{}lcccc@{}}
			\toprule
			Model / est.        & Task & \cellhead{Primary                                 \\repair} & \cellhead{Random repair\\min--max} & \cellhead{Random $\geq$\\primary} \\
			\midrule
			Qwen3-0.6B / $\phi$ & MLX  & $0.005$           & $-0.038$ to $-0.017$ & $0/10$ \\
			Qwen3-0.6B / $\phi$ & FWE  & $0.034$           & $-0.007$ to $0.021$  & $0/10$ \\
			Qwen3-0.6B / $\phi$ & MK2  & $0.144$           & $-0.024$ to $0.020$  & $0/10$ \\
			Qwen3-0.6B / RESA   & MLX  & $0.733$           & $0.106$ to $0.123$   & $0/10$ \\
			Qwen3-0.6B / RESA   & FWE  & $0.176$           & $0.045$ to $0.057$   & $0/10$ \\
			Qwen3-0.6B / RESA   & MK2  & $0.257$           & $-0.021$ to $0.016$  & $0/10$ \\
			Qwen3-1.7B / $\phi$ & MLX  & $0.099$           & $-0.023$ to $-0.004$ & $0/10$ \\
			Qwen3-1.7B / $\phi$ & FWE  & $0.362$           & $0.018$ to $0.094$   & $0/10$ \\
			Qwen3-1.7B / $\phi$ & MK2  & $0.083$           & $-0.016$ to $0.019$  & $0/10$ \\
			Qwen3-1.7B / RESA   & MLX  & $0.449$           & $0.072$ to $0.098$   & $0/10$ \\
			Qwen3-1.7B / RESA   & FWE  & $0.225$           & $0.040$ to $0.057$   & $0/10$ \\
			Qwen3-1.7B / RESA   & MK2  & $0.256$           & $0.117$ to $0.158$   & $0/10$ \\
			\bottomrule
		\end{tabular}}
	\printpapertable{tab:random}
\end{table}

\subsection{Generic-calibration stability}
Table~\ref{tab:generic-calibration} reports selected-set stability across balanced 3/5/10-sequence-per-corpus prefixes, corpus-specific estimates, and leave-one-corpus-out comparisons. Table~\ref{tab:g-robustness} uses alternative summaries of the same calibration rows. The frozen masking rule continues to use mean generic local reconstruction gain; alternative summaries do not redefine the mask. All generic-calibration observations satisfy the residual-energy validity floor.
\begin{table}[t]
	\centering
	\small
	\setlength{\tabcolsep}{4.0pt}
	\renewcommand{\arraystretch}{1.12}
	\caption{Generic-calibration stability. Robust counts refer to layers negative by confidence interval, on all three corpora, and in every leave-one-corpus-out analysis (LOCO). Prefix agreement compares the selected sign pattern with the full calibration.}
	\label{tab:generic-calibration}
	\sbox{\papertablebox}{%
		\begin{tabular}{@{}lcccc@{}}
			\toprule
			Model        & Est.   & $B$  & \cellhead{Robust negative counts                         \\CI/corpus/LOCO} & \cellhead{Prefix agreement\\3/5/10 per corpus} \\
			\midrule
			Llama-3.2-1B & $\phi$ & $0$  & 0 / 0 / 0                        & 1.000 / 1.000 / 1.000 \\
			Llama-3.2-1B & RESA   & $0$  & 0 / 0 / 0                        & 1.000 / 1.000 / 1.000 \\
			Llama-3.2-3B & $\phi$ & $0$  & 0 / 0 / 0                        & 1.000 / 1.000 / 1.000 \\
			Llama-3.2-3B & RESA   & $0$  & 0 / 0 / 0                        & 1.000 / 1.000 / 1.000 \\
			Qwen3-0.6B   & $\phi$ & $9$  & 8 / 8 / 9                        & 1.000 / 1.000 / 1.000 \\
			Qwen3-0.6B   & RESA   & $12$ & 11 / 10 / 11                     & 0.964 / 1.000 / 1.000 \\
			Qwen3-1.7B   & $\phi$ & $5$  & 4 / 5 / 5                        & 1.000 / 1.000 / 1.000 \\
			Qwen3-1.7B   & RESA   & $12$ & 8 / 7 / 9                        & 0.893 / 0.964 / 1.000 \\
			\bottomrule
		\end{tabular}}
	\printpapertable{tab:generic-calibration}
\end{table}

\begin{table}[t]
	\centering
	\small
	\setlength{\tabcolsep}{4.0pt}
	\renewcommand{\arraystretch}{1.12}
	\caption{Sensitivity of the generic negative-$G$ layer set. $B$ is the selected count. Each pair gives layer-sign agreement and negative-set Jaccard similarity to the mean-$G$ masking rule. Every calibration observation passes the validity floor.}
	\label{tab:g-robustness}
	\sbox{\papertablebox}{%
		\begin{tabular}{@{}lcccc@{}}
			\toprule
			Model / est.          & $B$  & \cellhead{Median                                     \\agreement/Jaccard} & \cellhead{Trimmed\\agreement/Jaccard} & \cellhead{Energy-pooled\\agreement/Jaccard} \\
			\midrule
			Llama-3.2-1B / $\phi$ & $0$  & $1.0 / 1.0$      & $1.0 / 1.0$     & $1.0 / 1.0$     \\
			Llama-3.2-1B / RESA   & $0$  & $1.0 / 1.0$      & $1.0 / 1.0$     & $1.0 / 1.0$     \\
			Llama-3.2-3B / $\phi$ & $0$  & $1.0 / 1.0$      & $1.0 / 1.0$     & $1.0 / 1.0$     \\
			Llama-3.2-3B / RESA   & $0$  & $1.0 / 1.0$      & $1.0 / 1.0$     & $1.0 / 1.0$     \\
			Qwen3-0.6B / $\phi$   & $9$  & $0.964 / 0.889$  & $1.0 / 1.0$     & $0.964 / 0.889$ \\
			Qwen3-0.6B / RESA     & $12$ & $0.929 / 0.833$  & $0.964 / 0.917$ & $0.964 / 0.917$ \\
			Qwen3-1.7B / $\phi$   & $5$  & $1.0 / 1.0$      & $1.0 / 1.0$     & $0.964 / 0.8$   \\
			Qwen3-1.7B / RESA     & $12$ & $0.857 / 0.667$  & $0.893 / 0.75$  & $0.929 / 0.833$ \\
			\bottomrule
		\end{tabular}}
	\printpapertable{tab:g-robustness}
\end{table}

\clearpage
\section{Estimator Dependence and Auxiliary Representations}
\label{app:pisa}
\subsection{\PISAZero masking}
We apply the same negative-$G$ masking rule to a zeroth-order estimator adapted from PISA, whose original formulation concerns diffusion transformers~\citep{li2026pisa}. The omitted middle region is divided into 64-token blocks, with one zeroth-order summary per block alongside the exact selected-token branch. Across 12 model--task conditions, fidelity repair is positive in 2, unresolved in 5, and negative in 5; every utility interval is unresolved (Table~\ref{tab:pisa-condition}). This experiment serves as an estimator-family transfer test.

\begin{table}[t]
	\centering
	\small
	\setlength{\tabcolsep}{4.0pt}
	\renewcommand{\arraystretch}{1.12}
	\caption{\PISAZero condition-level masking effects relative to unmodified \PISAZero. MLX: Multi-LexSum; MK2: NIAH Multikey-2. The table reports all effects, including degradations.}
	\label{tab:pisa-condition}
	\sbox{\papertablebox}{%
		\begin{tabular}{@{}lcc@{}}
			\toprule
			Task & Fidelity repair                & Utility repair                \\
			\midrule
			\multicolumn{3}{@{}l}{\textit{Llama-3.2-1B}}                          \\
			FWE  & \estci{-0.005}{-0.016}{0.005}  & \estci{-0.040}{-0.093}{0.007} \\
			MLX  & \estci{-0.065}{-0.077}{-0.054} & \estci{3.818}{-1.308}{9.062}  \\
			MK2  & \estci{-0.033}{-0.068}{-0.004} & \estci{-0.040}{-0.100}{0.000} \\
			\addlinespace[3pt]
			\multicolumn{3}{@{}l}{\textit{Llama-3.2-3B}}                          \\
			FWE  & \estci{0.006}{0.001}{0.011}    & \estci{0.027}{0.000}{0.060}   \\
			MLX  & \estci{-0.001}{-0.002}{0.0004} & \estci{2.228}{-2.750}{6.860}  \\
			MK2  & \estci{0.007}{0.001}{0.013}    & \estci{0.000}{-0.060}{0.060}  \\
			\addlinespace[3pt]
			\multicolumn{3}{@{}l}{\textit{Qwen3-0.6B}}                            \\
			FWE  & \estci{-0.003}{-0.005}{-0.001} & \estci{0.013}{-0.033}{0.060}  \\
			MLX  & \estci{-0.002}{-0.004}{-0.001} & \estci{-1.191}{-5.567}{3.619} \\
			MK2  & \estci{-0.002}{-0.005}{0.0001} & \estci{-0.040}{-0.100}{0.000} \\
			\addlinespace[3pt]
			\multicolumn{3}{@{}l}{\textit{Qwen3-1.7B}}                            \\
			FWE  & \estci{-0.003}{-0.008}{0.002}  & \estci{0.000}{0.000}{0.000}   \\
			MLX  & \estci{-0.004}{-0.007}{-0.001} & \estci{1.216}{-3.221}{6.067}  \\
			MK2  & \estci{0.002}{-0.001}{0.005}   & \estci{-0.020}{-0.060}{0.000} \\
			\bottomrule
		\end{tabular}}
	\printpapertable{tab:pisa-condition}
\end{table}

\clearpage
\section{Broader Model Results}
\label{app:stage-endpoints}
\subsection{All-layer completion endpoints}
Here, \emph{all-layer completion} denotes the unmodified completion configuration before negative-$G$ masking. Tables~\ref{tab:llama-stage} and~\ref{tab:qwen-stage} show local reconstruction gain, final-fidelity gain, and free-running task-utility gain for this configuration on the same $N=50$ mechanism request sets. Utility is the request-paired method-minus-\ExactTopK gain. All 12 Llama conditions have CI-positive local reconstruction and final-fidelity gains; eight Qwen conditions combine CI-positive post-$W_O$ local reconstruction gain with CI-negative final-fidelity gain.
\begin{table}[t]
	\centering
	\small
	\setlength{\tabcolsep}{4.0pt}
	\renewcommand{\arraystretch}{1.12}
	\caption{Llama all-layer completion endpoints. MLX denotes Multi-LexSum and MK2 denotes NIAH Multikey-2. Utility is the request-paired method-minus-\ExactTopK gain on the same N=50 mechanism request set. These are observational comparisons of all-layer completion, not the single-layer direct-runtime experiment. Values are means and paired 95\% intervals.}
	\label{tab:llama-stage}
	\sbox{\papertablebox}{%
		\begin{tabular}{@{}lcccc@{}}
			\toprule
			Task & Est.   & Local $G$                   & Final-fidelity $F$          & Utility gain $U$ vs. \ExactTopK \\
			\midrule
			\multicolumn{5}{@{}l}{\textit{Llama-3.2-1B}}                                                                \\
			MLX  & $\phi$ & \estci{0.473}{0.464}{0.482} & \estci{0.333}{0.294}{0.371} & \estci{9.174}{3.690}{14.497}    \\
			FWE  & $\phi$ & \estci{0.438}{0.421}{0.454} & \estci{0.176}{0.138}{0.215} & \estci{0.307}{0.233}{0.380}     \\
			MK2  & $\phi$ & \estci{0.457}{0.450}{0.463} & \estci{0.456}{0.333}{0.590} & \estci{0.360}{0.200}{0.520}     \\
			MLX  & RESA   & \estci{0.300}{0.291}{0.310} & \estci{0.275}{0.242}{0.308} & \estci{16.653}{11.451}{22.198}  \\
			FWE  & RESA   & \estci{0.544}{0.530}{0.557} & \estci{0.148}{0.114}{0.182} & \estci{0.360}{0.293}{0.427}     \\
			MK2  & RESA   & \estci{0.374}{0.368}{0.379} & \estci{0.366}{0.253}{0.492} & \estci{0.260}{0.100}{0.420}     \\
			\addlinespace[3pt]
			\multicolumn{5}{@{}l}{\textit{Llama-3.2-3B}}                                                                \\
			MLX  & $\phi$ & \estci{0.516}{0.509}{0.523} & \estci{0.137}{0.121}{0.155} & \estci{3.135}{-2.251}{8.971}    \\
			FWE  & $\phi$ & \estci{0.620}{0.606}{0.633} & \estci{0.082}{0.050}{0.116} & \estci{0.113}{0.013}{0.213}     \\
			MK2  & $\phi$ & \estci{0.612}{0.599}{0.624} & \estci{0.105}{0.068}{0.142} & \estci{0.160}{0.020}{0.320}     \\
			MLX  & RESA   & \estci{0.374}{0.364}{0.385} & \estci{0.117}{0.099}{0.136} & \estci{8.128}{2.123}{14.702}    \\
			FWE  & RESA   & \estci{0.652}{0.641}{0.661} & \estci{0.037}{0.009}{0.064} & \estci{0.153}{0.073}{0.240}     \\
			MK2  & RESA   & \estci{0.604}{0.583}{0.622} & \estci{0.147}{0.103}{0.188} & \estci{0.140}{0.000}{0.280}     \\
			\bottomrule
		\end{tabular}}
	\printpapertable{tab:llama-stage}
\end{table}

\begin{table}[t]
	\centering
	\small
	\setlength{\tabcolsep}{4.0pt}
	\renewcommand{\arraystretch}{1.12}
	\caption{Qwen all-layer completion endpoints. MLX denotes Multi-LexSum and MK2 denotes NIAH Multikey-2. Utility is the request-paired method-minus-\ExactTopK gain on the same N=50 mechanism request set. These are observational comparisons of all-layer completion, not the single-layer direct-runtime experiment. Values are means and paired 95\% intervals.}
	\label{tab:qwen-stage}
	\sbox{\papertablebox}{%
		\begin{tabular}{@{}lcccc@{}}
			\toprule
			Task & Est.   & Local $G$                      & Final-fidelity $F$             & Utility gain $U$ vs. \ExactTopK   \\
			\midrule
			\multicolumn{5}{@{}l}{\textit{Qwen3-0.6B}}                                                                          \\
			MLX  & $\phi$ & \estci{0.322}{0.314}{0.331}    & \estci{0.102}{0.076}{0.129}    & \estci{0.620}{-5.626}{6.219}      \\
			FWE  & $\phi$ & \estci{-0.157}{-0.401}{0.015}  & \estci{-0.035}{-0.049}{-0.022} & \estci{0.100}{0.020}{0.180}       \\
			MK2  & $\phi$ & \estci{0.077}{0.054}{0.098}    & \estci{-0.105}{-0.133}{-0.080} & \estci{-0.180}{-0.300}{-0.080}    \\
			MLX  & RESA   & \estci{0.037}{0.031}{0.044}    & \estci{-0.809}{-0.907}{-0.718} & \estci{-18.532}{-24.227}{-12.824} \\
			FWE  & RESA   & \estci{0.096}{0.067}{0.125}    & \estci{-0.204}{-0.242}{-0.168} & \estci{-0.027}{-0.100}{0.053}     \\
			MK2  & RESA   & \estci{0.050}{0.047}{0.053}    & \estci{-0.269}{-0.321}{-0.222} & \estci{-0.200}{-0.320}{-0.100}    \\
			\addlinespace[3pt]
			\multicolumn{5}{@{}l}{\textit{Qwen3-1.7B}}                                                                          \\
			MLX  & $\phi$ & \estci{0.227}{0.210}{0.241}    & \estci{0.007}{-0.026}{0.040}   & \estci{-2.550}{-7.428}{2.177}     \\
			FWE  & $\phi$ & \estci{-0.250}{-0.338}{-0.165} & \estci{-0.353}{-0.437}{-0.272} & \estci{-0.133}{-0.213}{-0.053}    \\
			MK2  & $\phi$ & \estci{0.101}{0.085}{0.117}    & \estci{-0.123}{-0.148}{-0.099} & \estci{-0.460}{-0.600}{-0.320}    \\
			MLX  & RESA   & \estci{0.042}{0.037}{0.046}    & \estci{-0.492}{-0.548}{-0.440} & \estci{-12.023}{-17.141}{-6.779}  \\
			FWE  & RESA   & \estci{0.093}{0.083}{0.102}    & \estci{-0.245}{-0.295}{-0.197} & \estci{-0.127}{-0.207}{-0.047}    \\
			MK2  & RESA   & \estci{0.125}{0.119}{0.130}    & \estci{-0.285}{-0.319}{-0.252} & \estci{-0.460}{-0.600}{-0.320}    \\
			\bottomrule
		\end{tabular}}
	\printpapertable{tab:qwen-stage}
\end{table}

\subsection{Focused Qwen3-8B result}
\label{app:scale8b}
A focused RESA experiment uses Qwen3-8B layers L6, L17, and L29, selected from layers with positive generic-calibration local reconstruction gain $G$ before downstream evaluation and crossed with three tasks at 32K native and 64K YaRN2 (18 actions). Final-fidelity gain is CI-positive in 7 actions, unresolved in 10, and CI-negative in 1. The sole CI-negative action, native-32K Multi-LexSum/RESA L6, is characterized post hoc: captured-state local reconstruction gain $G=0.393$ $[0.349,0.422]$, selected-block $G_{\mathrm{block}}=0.057$ $[0.043,0.067]$, first CI-negative \Full-relative block at 29, and final-fidelity gain $F=-0.0024$ $[-0.0040,-0.0008]$.

\end{document}